\documentclass[pdflatex,sn-aps]{sn-jnl}

\normalbaroutside
\usepackage{booktabs}
\usepackage{amsmath,amssymb,amsfonts}
\usepackage{tikz}
\usetikzlibrary{chains,
                fit,
                positioning,
                shapes,
                shadows,
                trees,
                patterns,
                matrix}
\usepackage{pgfplots}
\pgfdeclarelayer{background}
\pgfdeclarelayer{foreground}
\pgfsetlayers{background,main,foreground}
\pgfplotsset{compat=newest}
\usepackage{program}
\usepackage{pgfplotstable}
\usepgfplotslibrary{colorbrewer}
\pgfplotsset{cycle list/Set1}
\pgfplotsset{
    discard if not/.style 2 args={
        x filter/.code={
            \edef\tempa{\thisrow{#1}}
            \edef\tempb{#2}
            \ifx\tempa\tempb
            \else
                
            \fi
        }
    }
} 

\usepackage[algo2e, linesnumbered,ruled,vlined]{algorithm2e} 
\SetCommentSty{mycommfont}

\usepackage{pgfplotstable}
\usepackage[flushleft]{threeparttable}
\usepackage{subcaption}

\usepackage{makecell}
\usepackage{color}
\usepackage{colortbl}
\newcommand{\rw}[1]{\cellcolor{red!20}{#1}}
\newcommand{\rf}[1]{\cellcolor{black!20}{#1}}
\renewcommand{\bf}[1]{\cellcolor{black!20}\textbf{#1}}

\usepackage[acronym]{glossaries}
\glsdisablehyper
\newacronym{AGSR}{AGSR}{Average Generations to Success Rate}
\newacronym{AESR}{AESR}{Average Evaluations to Success Rate}
\newacronym{EA}{EA}{Evolutionary Algorithm}
\newacronym{MPO}{MPO}{Multi-period Portfolio Optimisation}
\newacronym{SPO}{SPO}{Single-period Portfolio Optimisation}
\newacronym{DP}{DP}{Dynamic Programming}
\newacronym{SP}{SP}{Stochastic Programming}
\newacronym{DCP}{DCP}{Disciplined Convex Programming}
\newacronym{MOP}{MOP}{Multi-Objective optimisation Problem}
\newacronym{MO}{MO}{Multi-Objective}
\newacronym{ParDen-Sur}{ParDen-Sur}{Pareto Driven Surrogate}
\newacronym{SR}{SR}{Success Rate}
\newacronym{MDP}{MDP}{Markov Decision Problem}

\usepackage{siunitx}

\jyear{2022}%

\theoremstyle{thmstyleone}%
\theoremstyle{thmstyletwo}%

\theoremstyle{thmstylethree}%

\begin{document}

\title[Article Title]{Pareto Driven Surrogate (ParDen-Sur) Assisted Optimisation of Multi-period Portfolio Backtest Simulations}


\author*[1]{\fnm{Terence L.} \sur{van Zyl}}\email{tvanzyl@gmail.com}

\author[2]{\fnm{Matthew} \sur{Woolway}}\email{matt.woolway@gmail.com}

\author[3]{\fnm{Andrew} \sur{Paskaramoorthy}}\email{ab.paskaramoorthy@gmail.com}

\affil*[1]{\orgdiv{Institute for Intelligent Systems}, \orgname{University of Johannesburg}, \orgaddress{\city{Johannesburg}, \country{South Africa}}}

\affil[2]{\orgdiv{Faculty of Engineering and the Built Environment}, \orgname{University of Johannesburg}, \orgaddress{\city{Johannesburg}, \country{South Africa}}}

\affil[3]{\orgdiv{Department of Statistical Sciences}, \orgname{University of Cape Town}, \orgaddress{\city{Cape Town}, \country{South Africa}}}

\abstract{
Portfolio management is a multi-period multi-objective
optimisation problem subject to a wide range of constraints. 
However, in practice, portfolio management is treated as a single-period problem partly due to the computationally burdensome hyper-parameter search procedure needed to construct a multi-period Pareto frontier.  
This study presents the \gls{ParDen-Sur} modelling framework to efficiently perform the required hyper-parameter search. \gls{ParDen-Sur} extends previous surrogate frameworks by including a reservoir sampling-based look-ahead mechanism for offspring generation in \glspl{EA} alongside the traditional acceptance sampling scheme. We evaluate this framework against, and in conjunction with, several seminal \gls{MO} \glspl{EA} on two datasets for both the single- and multi-period use cases. 
Our results show that \gls{ParDen-Sur} can speed up the exploration for optimal hyper-parameters by almost $2\times$ with a statistically significant improvement of the Pareto frontiers, across multiple \glspl{EA}, for both datasets and use cases.


}

\keywords{portfolio optimisation, surrogate modelling, multi objective optimisation, evolutionary algorithm, artificial intelligence, backtesting, hyper-parameter selection}



\maketitle
\section{Introduction}\label{sec:introduction}\glsresetall
Markowitz \cite{art:markowitz1952} formulated the portfolio selection problem as a \gls{MOP}, where an investor aims to satisfy the conflicting objectives of maximising return whilst minimising risk. Since this seminal work, the portfolio optimisation problem has been extended to include a range of additional constraints and objectives to reflect real-world considerations \cite{boyd2017multi}. The competing objectives are typically combined linearly into a single function using trade-off parameters that reflect the relative importance of each objective (i.e. they ``trade-off" different objectives). By varying the risk-aversion parameter and solving the optimisation problem, one can uncover the set of non-dominated solutions known as the Pareto frontier or, in the finance jargon, the efficient frontier\footnote{We use the terms Pareto frontier and efficient frontier interchangeably throughout this work}~\cite{paskaramoorthy2020framework,paskaramoorthy2021efficient}. However, not all choices of trade-off parameters yield non-dominated solutions in the risk-return space, resulting in unnecessary computation. Thus, the key issue is determining the efficient frontier with as few evaluations as possible.

\gls{MO} \glspl{EA} have been extensively applied in portfolio optimisation~\cite{art:ponsich2012survey}. However, portfolio optimisation is treated as a deterministic problem in these studies. In reality, the risk and return terms of the objective function are estimated from data and are hence stochastic. In the finance literature, the objective value of the optimised portfolio is considered an ``in-sample", and hence a highly inflated estimate of true performance~\cite{art:kan2008distribution}. Consequently, it is common practice that portfolio performance is estimated on separate test data using a process known as ``\textit{backtesting}". This process is computationally burdensome since it involves repeatedly solving an optimisation problem at each time step in the backtest period for each choice of trade-off parameters~\cite{van2021parden}.

In practice, portfolio management involves making a sequence of portfolio decisions over time and is thus more accurately represented as a multi-period optimisation problem~\cite{book:lee2000theory}. Taking a multi-period perspective of the portfolio optimisation problem confers several advantages over the single-period counterpart. For example, \gls{MPO} allows for management of trading costs \cite{hendricks2014reinforcement}, time-varying forecasts, intertemporal hedging, and other intertemporal constraints~\cite{kolm201460}. However, realising these advantages requires that some challenges are overcome. Firstly, \gls{MPO} requires multi-period forecasts adding difficulty to the already challenging task of single-period forecasting. Secondly, \gls{MPO} is computationally burdensome, especially when incorporating real-world considerations. Consequently, portfolio management in practice typically consists of repeatedly performing myopic single-period optimisations.

Traditionally portfolio optimisation is extended to a multi-period setting using stochastic optimal control as a framework and solved using \gls{DP}. However, formulations relying on \gls{DP} are typically unable to adequately capture real-world market dynamics and investment considerations primarily due to the ``\textit{curse of dimensionality}". Instead, approximate policies can be determined by considering simpler (although sub-optimal) \gls{SP} formulations that do not depend on Bellman Optimality. For example, the single-period mean-variance problem can be extended to the multi-period case by including additional risk and return objectives and additional constraints for each period~\cite{boyd2017multi}. However, this extension negatively impacts the number of decision variables, and hence the computational complexity can grow exponentially with the investment horizon.

This work addresses the computational difficulties associated with backtesting when determining the efficient frontier during a realistic \gls{MPO} problem. Following Boyd \textit{et al.}~\cite{boyd2017multi}, this study adopts the stochastic perspective of the portfolio optimisation problem and treats trade-off parameters as hyper-parameters to be optimised using backtest performance estimates. Nystrup \textit{et al.}~\cite{nystrup2020hyperparameter} shows that using the MO-CMA-ES \gls{EA} for determining the efficient frontier improves upon the grid search procedure used by Boyd \textit{et al.}~\cite{boyd2017multi}.  Most recently, van Zyl \textit{et al.}~\cite{van2021parden} compared several seminal \gls{MO} \glspl{EA} and proposed a procedure that includes a surrogate model, which further outperforms existing procedures. 

This work extends that of van Zyl \textit{et al.}~\cite{van2021parden} by investigating \gls{MPO}. Most existing work in \gls{MPO} tackles the problem of determining an approximately optimal policy, treating \gls{MPO} as a single objective problem. Consideration of the efficient frontier is limited to the case when the problem is specified as a \gls{MO} problem and an analytical solution is available (for example, \cite{zhou2000continuous, li2000optimal}). In contrast, this work considers the problem of trying to efficiently determine the Pareto frontier for a more realistic portfolio optimisation problem containing several constraints. The main aim of which is to improve not only the efficiency of the algorithms but also the overall quality and performance of the solutions. To this end, the contributions of this study are:
\begin{itemize}
    \item an empirical performance comparison of several seminal \gls{MOP} \gls{EA} for both \gls{SPO} and \gls{MPO},
    \item the demonstration of the superior efficacy of surrogate assisted optimisation to alleviate the computational burden of constructing the Pareto frontier, 
    \item and the presentation of a new algorithm, \gls{ParDen-Sur}, which improves upon state-of-the-art in surrogate assisted portfolio optimisation by introducing a reservoir sampling based look-ahead mechanism.
\end{itemize}

\section{Background}\label{sec:background}

\subsection{Mean-Variance Portfolio Optimisation}
The original mean-variance optimisation problem is specified by a bi-objective criterion, where risk is measured by the variance of the portfolio's return \cite{art:markowitz1952}. These objectives are combined linearly into a single optimisation problem:
\begin{eqnarray}
    &\text{maximise} \quad &\mathbf{w}'\mathbf{\mu} - \frac{\gamma}{2}\mathbf{w}'\Sigma \mathbf{w} \\
    &\text{subject to} \quad &\mathbf{w}'\mathbf{1} = 1
\end{eqnarray}
where $\mathbf{w}$ is the vector of portfolio weights, $\mathbf{\mu}$ is the vector of expected returns, $\Sigma$ is the covariance matrix of returns, and $\gamma$ is known as the \textit{risk-aversion} parameter, which is used to trade-off the conflicting objectives. The resulting problem is a quadratic program that has an analytical solution. 

A particular choice of the risk-aversion parameter yields a particular optimal solution. The efficient frontier can be determined by varying the trade-off parameters and solving the optimisation problem. It is often of interest to determine the portfolio from the efficient frontier that maximises the Sharpe Ratio, defined as the average return (in excess of the return on a risk-free asset) of the portfolio over the standard deviation of return. In this case, the optimisation procedure comprises two steps: first, determine the efficient frontier, and second, select the Sharpe-ratio maximising portfolio.

\glsreset{MPO}
\subsection{Approaches to \gls{MPO}}

Starting with Merton \cite{merton1969lifetime, merton1975optimum}, and others \cite{art:mossin1968optimal, art:samuelson1975lifetime,art:constantinides1979multiperiod}, \gls{MPO} has been studied as a \gls{MDP} where optimal policies are found using \gls{DP}. Here, the portfolio optimisation problem is typically restricted to idealised settings to enable the construction and solution of the Bellman optimality equation. In particular, it is typical for the state-space to be limited to very few variables to avoid the curse of dimensionality and for the state transition function to be treated as known. In addition, the objective function is often chosen from a class of utility functions with convenient analytic properties, but which may render the problem as a single-objective optimisation problem where the risk-return trade-off is not made explicit. Lastly, problems are restricted to one or two constraints of theoretical interest. Thus, studies investigating \gls{MPO} through \gls{DP} are largely theoretical in nature, focusing on qualitative understanding of normative investor behaviour.

In contrast, \gls{SP} is a highly flexible modelling framework for multi-period optimisation, which can incorporate a diverse array of real-world considerations, including alternative risk measures, risks and uncertainty hedging, hard and soft constraints, and intermediate goals (including stochastic and irregular outflows)~\cite{boyd2017multi}. Furthermore, \glspl{SP} model uncertainty using scenarios, defined as sequences of states, and hence are not restricted to \gls{MDP} formulations. However, a \gls{SP} becomes computationally intractable as its specification becomes extensive, and instead, an approximate program is used to find a policy. Common approximations include discretisation of the observed variables, aggregation of stages, reduction of the problem horizon, and replacing random quantities with their expectations~\cite{art:powell2019unified}. When the data's expectations are unknown, they are replaced with forecasts that are sequentially updated at each timestep, more broadly referred to as a rolling horizon procedure or Model Predictive Control (MPC) and widely adopted in practice. 

\subsection{A Convex Optimisation Framework for Portfolio Optimisation}

Using the \gls{SP} approach, Boyd \textit{et al.} \cite{boyd2017multi} presents a convex optimisation framework for \gls{MPO}, which can be used for MPC and includes the single-period problem as a special case. Notably, trading and holding costs are treated as soft constraints selected to achieve some desired trading behaviour and are typically used as regularisation methods. The portfolio optimisation problem can be equivalently formulated by including the soft constraints as additional criteria in the objective function, with associated trade-off parameters:
\begin{align}\label{eqn:cvxpo}
     & \text{maximise} \quad \sum_{\tau=t+1}^{t+H} \mathbf{w}_{\tau}'\mathbf{\mu}_{\tau} - \frac{\gamma}{2}\mathbf{w}_{\tau}'\Sigma_{\tau} \mathbf{w}_{\tau} -  \gamma_{\text{trade}}\phi^\text{trade}\left(\mathbf{w}_{\tau} - \mathbf{w}_{\tau-1}\right) \nonumber \\
     & \hspace{3cm}- \gamma_{\text{hold}} \phi^\text{hold}\left(\mathbf{w}_{\tau}\right) \nonumber                                                                                                                                                                   \\
     & \text{subject to} \quad \mathbf{w}_{\tau}'\mathbf{1} = 1, \mathbf{w}_{\tau} - \mathbf{w}_{\tau-1} \in \mathcal{Z}_{\tau}, \mathbf{w}_{\tau} \in \mathcal{W}_{\tau}
\end{align}
where $H$ is a natural number representing the investment horizon, $\mathbf{\mu}_{\tau}$ and $\Sigma_{\tau}$ are the return and risk forecasts for each period $\tau = t, \dots, t+H$, $\mathcal{W}_{\tau}$ and $\mathcal{Z}_{\tau}$ are convex sets representing the holding and trading constraints respectively, and $\gamma$, $\gamma_{\text{trade}}$, and $\gamma_{\text{hold}}$ are the trade-off parameters for risk, trading costs, and holding costs respectively. Note that setting $H=1$ specifies the single-period problem.

The extended formulation of the mean-variance problem in Equation \ref{eqn:cvxpo} is analytically intractable. Computational methods determine the efficient frontier by repeatedly solving the optimisation problem for different trade-off parameters, but, unlike Markowitz's original formulation, not all selected trade-off parameters are associated with Pareto optimal solutions, resulting in unnecessary computation. This procedure of repeatedly optimising for different trade-off parameters becomes increasingly costly as the number of decision variables grows, thus limiting the investment horizon.

\glsreset{EA}
\subsection{\glspl{EA}}\label{sec:metaheuristic}


There exists numerous evolutionary \gls{MOP} algorithms in the literature~\cite{art:liu2020multi,bowditch2019comparative}. Amongst these, certain algorithms have been shown to be successful in applications across several different domains~\cite{art:deb2002fast,art:zhang2007moea,art:igel2007covariance,art:helbig2014population,art:alsattar2020mogsabat}. The use of \glspl{EA} for finding solutions to portfolio optimisation problems has been covered extensively in the portfolio optimisation literature \cite{art:doering2019metaheuristics,art:loukeris2009numerical,art:fernandez2015hybrid} with focus almost exclusively given to the single-period problem. The presented study aims to apply various representative \glspl{EA} and demonstrate the utility of a surrogate-assisted evolutionary optimisation approach to solving portfolio optimisation problems described in this paper. To this end, this study implements the evolutionary \gls{MOP} methods described below.

\subsubsection{MO-CMA-ES}

The Covariance Matrix Adaptation Evolution Strategy (CMA-ES) is an elitist evolutionary strategy for numerical optimisation. Multi-objective CMAES (MO-CMA-ES), maintains a population of individuals that adapt their search strategy as in the elitist CMA-ES. The elites are subjected to \gls{MO} selection pressure. The selection pressure originates from non-dominated sorting using the crowding-distance first and then hypervolume as the second criterion \cite{art:igel2007covariance}.

\subsubsection{NSGA-II}

The NSGA-II algorithm follows the form of a Genetic Algorithm (GA) with modified crossover and survival. The surviving individuals are chosen Pareto front-wise first. There arises a situation where a front needs to be split since not all individuals can survive. When splitting the front, solutions are selected based on a Manhattan crowding distance in the objective space. However, the method seeks to hold onto the extreme points from each generation, and as a result, they are assigned a crowding distance of infinity. Furthermore, the method uses a binary tournament mating selection to increase selection pressure. Each individual is first compared by rank and then crowding distance~\cite{blank2020pymoo}. 

\subsubsection{R-NSGA-II}

The R-NSGA-II algorithm follows NSGA-II with a modified survival selection mechanism. Unlike NSGA-II, solutions are selected based on rank in splitting the front. This rank is calculated based on the Euclidean distance to a set of supplied reference points. The solution closest to a reference point is assigned a rank of one. The algorithm then selects the points with the highest rank to each reference point as offspring. After each reference point has selected the solution with the best rank for survival, all solutions within the epsilon distance of surviving solutions are ``cleared'', meaning they can not be selected for survival until all fronts have been processed. The cleared points are considered if more solutions still need to be selected. The free parameter epsilon has the effect that a smaller value results in a tighter set of solutions~\cite{blank2020pymoo}.

\subsubsection{NSGA-III}

The NSGA-III algorithm is based on reference directions instead of reference points. For the survival mechanism, after non-dominated sorting, the algorithm has a modified approach to dealing with the front splitting. The algorithm fills up the underrepresented reference direction first. If the reference direction does not have any solution assigned, then the solution with the smallest perpendicular distance in the normalised objective space is assigned as the survivor. In the case that additional solutions for a reference direction are required, these are assigned randomly. Consequently, when this algorithm converges, each reference direction attempts to find a suitable non-dominated solution~\cite{blank2020pymoo}.

\subsubsection{R-NSGA-III} 

The R-NSGA-III algorithm follows the general NSGA-III procedure with a modified survival selection operator. First, non-dominated sorting is done as in NSGA-III. Solutions are associated with aspiration points based on perpendicular distance, and then solutions from the underrepresented reference direction are chosen. For this reason, when this algorithm converges, each reference line seeks to find a good representative, non-dominated solution~\cite{blank2020pymoo}.

\subsubsection{U-NSGA-III} 

It has previously been shown that tournament selection performs better than random selection. Unlike NSGA-III, which selects parents randomly for mating, the U-NSGA-III algorithm uses tournament pressure as an improvement~\cite{blank2020pymoo}.

\subsection{Backtesting}

Studies applying \glspl{EA} have largely treated the portfolio optimisation problem as a deterministic problem where the objective function is known. In reality, the objective function is unknown and is estimated from historical data. This reality has several important consequences for evaluating a portfolio, frontier, and solution method performance.

Firstly, the objective function value of the optimised portfolio is considered an ``in-sample'' and hence a highly inflated estimate of true performance. Consequently, it is common practice that portfolios are evaluated ``out-of-sample'', that is, using return data after the period of optimisation. When evaluating an entire policy, the portfolio at the beginning of each time period is evaluated using the returns of that period, commonly referred to as ``backtesting''. For policies constructed using a rolling window approach, a backtest necessarily involves an optimisation at each time period for each choice of trade-off parameters.

Secondly, there is no guarantee that in-sample Pareto optimal solutions are optimal out-of-sample, motivating the use of backtest estimates to determine the Pareto frontier. In particular, motivated by standard practice in machine learning, Boyd \textit{et al.}~\cite{boyd2017multi} considered the set of trade-off parameters as the hyper-parameters and the portfolio weights as the parameters in a hybrid approach. They construct the Pareto frontier through a grid-search over the hyper-parameters, which entails carrying out a backtest for every hyper-parameter in the set considered.

Thirdly, since backtest estimates are sample-specific, statistical comparison between solution methods needs to account for an additional source of randomness arising from sampling variation of the backtest data. This is in contrast to the typical use of statistical methods to compare algorithm performance, which only accounts for the variation due to the stochastic elements of the algorithm. However, proposing a statistical test that includes this additional source of variation is outside the scope of this study.


\subsection{Surrogate-Assisted Optimisation}


Surrogate models are used in optimisation problems where the objective function is unknown and costly to evaluate. In the surrogate-assisted approach, objective values (outputs) are first evaluated for a sample of different decision variable choices (inputs). Then, a supervised learning algorithm (the surrogate) is trained on the input-output pairs to estimate the objective function. The surrogate model can then be used to assist in numerically optimising the objective function. The surrogate model can also be sequentially improved by including input-output pairs for candidate solutions at each iteration as the optimisation algorithm proceeds~\cite{book:sobester2008engineering}. This study's unknown objective function is the out-of-sample performance over the hyper-parameter space of both \gls{SPO} and \gls{MPO}.

\subsection{Motivation and Problem Statement}\label{sec:prob_statement}

\begin{figure}
    \centering
    \begin{tikzpicture}
        \begin{axis}[xlabel=Risk \%,
                ylabel=Return \%,
                axis background/.style={fill=gray!7, draw=gray},
                width=\textwidth,
                height=.9*\axisdefaultheight,
                legend cell align={left},
                legend style={at={(0.99,0.05)},
                        anchor=south east,legend columns=1,font=\footnotesize}]
            \addplot+[Set1-A, thick, mark=*, mark size = 1.75, mark options={
                        draw = Set1-A!90,
                        fill = Set1-A!70,
                        fill opacity=0.6,
                        draw opacity=0.3,
                    }]
            table[
                    x=Risk,
                    y=Return,
                    col sep=comma,
                ]{spo_pareto_results_510_djia_plot_true.txt};
            \addlegendentry{SPO Boyd 510}            
            \addplot+[Set1-C, thick, mark=*, mark size = 1.75, mark options={
                        draw = Set1-C!90,
                        fill = Set1-C!70,
                        fill opacity=0.6,
                        draw opacity=0.3,
                    }]
            table[
                    x=Risk,
                    y=Return,
                    col sep=comma,
                ]{mpo_pareto_results_510_djia_plot_true.txt};
            \addlegendentry{MPO Boyd 510}
            \addplot+[Set1-B, thick, mark=*, mark size = 1.75, mark options={
                        draw = Set1-B!90,
                        fill = Set1-B!70,
                        fill opacity=0.6,
                        draw opacity=0.3,
                    }]
            table[
                    x=Risk,
                    y=Return,
                    col sep=comma,
                ]{spo_pareto_djia_results_all_plot_true.txt};
            \addlegendentry{SPO Boyd + Random 6510}
            \addplot+[Set1-D, thick, mark=*, mark size = 1.75, mark options={
                        draw = Set1-D!90,
                        fill = Set1-D!70,
                        fill opacity=0.6,
                        draw opacity=0.3,
                    }]
            table[
                    x=Risk,
                    y=Return,
                    col sep=comma,
                ]{mpo_pareto_djia_results_all_plot_true.txt};
            \addlegendentry{MPO Boyd + Random 6510}
egend         \end{axis}
    \end{tikzpicture}
    \caption{\gls{MPO} vs \gls{SPO} Motivating Example: Optimal Frontiers.}\label{fig:spo_mpo_motivating}
\end{figure}
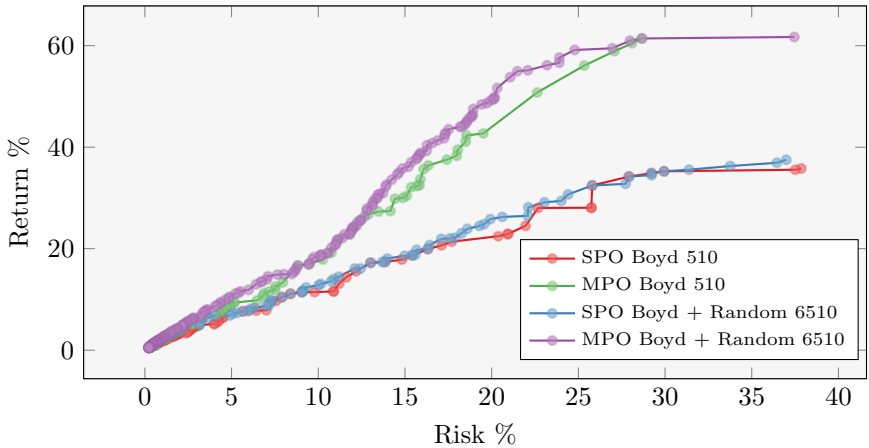

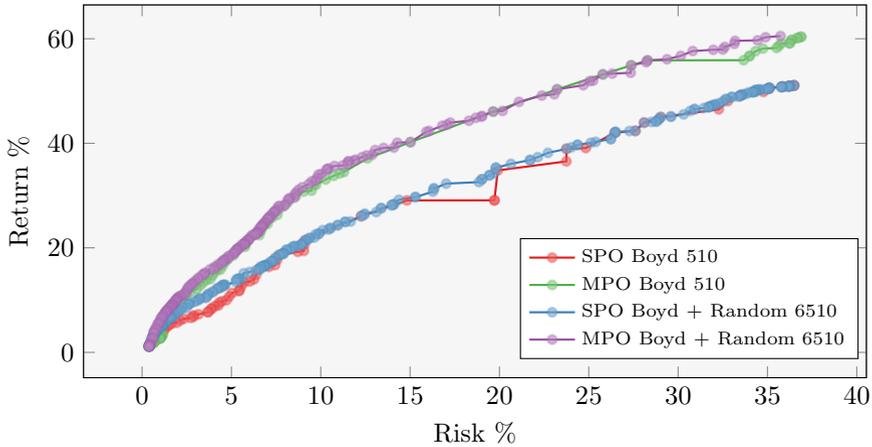
\begin{figure}
    \centering
    \begin{tikzpicture}
        \begin{axis}[xlabel=Risk \%,
                ylabel=Return \%,
                axis background/.style={fill=gray!7, draw=gray},
                width=\textwidth,
                height=.9*\axisdefaultheight,
                legend cell align={left},
                legend style={at={(0.99,0.05)},
                        anchor=south east,legend columns=1,font=\footnotesize}]
            \addplot+[Set1-A, thick, mark=*, mark size = 1.75, mark options={
                        draw = Set1-A!90,
                        fill = Set1-A!70,
                        fill opacity=0.6,
                        draw opacity=0.3,
                    }]
            table[
                    x=Risk,
                    y=Return,
                    col sep=comma,
                ]{spo_pareto_results_510_plot_true.txt};
            \addlegendentry{SPO Boyd 510}
            \addplot+[Set1-C, thick, mark=*, mark size = 1.75, mark options={
                        draw = Set1-C!90,
                        fill = Set1-C!70,
                        fill opacity=0.6,
                        draw opacity=0.3,
                    }]
            table[
                    x=Risk,
                    y=Return,
                    col sep=comma,
                ]{mpo_pareto_results_510_plot_true.txt};
            \addlegendentry{MPO Boyd 510}
            \addplot+[Set1-B, thick, mark=*, mark size = 1.75, mark options={
                        draw = Set1-B!90,
                        fill = Set1-B!70,
                        fill opacity=0.6,
                        draw opacity=0.3,
                    }]
            table[
                    x=Risk,
                    y=Return,
                    col sep=comma,
                ]{spo_pareto_results_all_plot_true.txt};
            \addlegendentry{SPO Boyd + Random 6510}
            \addplot+[Set1-D, thick, mark=*, mark size = 1.75, mark options={
                        draw = Set1-D!90,
                        fill = Set1-D!70,
                        fill opacity=0.6,
                        draw opacity=0.3,
                    }]
            table[
                    x=Risk,
                    y=Return,
                    col sep=comma,
                ]{mpo_pareto_results_all_plot_true.txt};
            \addlegendentry{MPO Boyd + Random 6510}
        \end{axis}
    \end{tikzpicture}
    \caption{\gls{SPO} vs \gls{MPO} Primary Example: Optimal Frontiers.}\label{fig:spo_mpo_primary}
\end{figure}

Real-world portfolio management can be described as a multi-period, \gls{MOP} subject to a potentially wide range of constraints. Here, the goal is to find a multi-period Pareto frontier comprised of policies with an optimal risk-return trade-off over the entire investment horizon.

However, due to analytical and computational intractability, approximate single-period frontiers are solved at each time step. The multi-period problem is reduced to a single-period problem for two reasons: multi-period forecasting is difficult, and the computational burden of multi-period optimisation is proportional to the number of periods considered.

In this study, the results deal solely with the latter problem, addressing the computational difficulty of finding a multi-period efficient frontier. Most previous studies applying \glspl{EA} do not recognise, firstly, that the Pareto frontier needs to be estimated on test data, and secondly, that most real-world considerations can be incorporated in a convex specification of the portfolio optimisation problem.

Recognising this, Boyd \textit{et al.}~\cite{boyd2017multi} and Nystrup \textit{et al.}~\cite{nystrup2020hyperparameter} adopt a hybrid approach, where the construction of the Pareto frontier is split into a hyper-parameter selection problem and a convex portfolio optimisation problem. However, a poor strategy for selecting candidate hyper-parameters can yield many dominated solutions, resulting in unnecessary computation.


In response, this study adopts the hybrid methodology of Boyd \textit{et al.}~\cite{boyd2017multi} and proposes including a surrogate model to conduct the hyper-parameter search more effectively when constructing the Pareto frontier.

\glsreset{ParDen-Sur}
\section{\gls{ParDen-Sur}}\label{sec:pareto_surrogate}

Surrogate methods have been extensively applied in single objective optimisation~\cite{art:ben2017universal,art:zhou2006combining,art:wan2005simulation,stander2020extended,perumal2020surrogate,stander2022surrogate,perumal2021surrogate,timilehin2021surrogate}, whereas the applications of surrogate assisted methods within a \gls{MO} setting are substantially less common. In particular, the literature for data-driven evolutionary \gls{MO} surrogate assisted optimisation is especially sparse~\cite{art:wang2018random,art:yang2019offline,art:chugh2017data,art:stander2020data}. To this end, the presented \gls{ParDen-Sur} framework is sufficiently general that it can be applied together with most evolutionary \gls{MO} algorithms. \gls{ParDen-Sur} also moves beyond generative models as surrogates, allowing for the integration of discriminative models~\cite{van2021parden}.

\gls{ParDen-Sur} relies on the use of supervised learning models to act as a surrogate drop-in replacement $\hat{f}(\cdot)$ to evaluate a computationally resource intensive real-world experiment or simulation $f(\cdot)$. The core idea of \gls{ParDen-Sur} is to limit the evaluation of candidates using the simulation $f(\cdot)$ to only those that have a high probability of advancing the Pareto front (the non-dominated solution set $\mathcal{P}$). The likelihood that a candidate would advance the Pareto front is determined by: i) whether the surrogate model predicts that the candidate advances the Pareto front and ii) the extent to which the surrogate is correct. In particular, \gls{ParDen-Sur} attempts to estimate the probability that the surrogate is correct, which is used as a threshold within an acceptance sampling scheme, and to set the size of a reservoir sampling look-ahead scheme when selecting candidates.

In order to determine the extent to which the surrogate is correct, a mechanism is required for assessing the probability $P\left[(x_i, y_i) \in \mathcal{P} | \hat{f}\right]$ that a pretender $(x_i,  y_i)$ will be in the current non-dominated set $\mathcal{P}$ given the surrogate's prediction $\hat{y}_i = \hat{f}(\cdot)$. \gls{ParDen-Sur} employs $k$-fold cross-validation $\operatorname{CV}_k(\cdot)$ together with the non-dominating rank~\cite{book:selvi2018application} of each candidate to estimate this probability as a non-dominated score ($\operatorname{NDScore}$).

Calculating an $\operatorname{NDScore}$ requires fitting surrogates $\hat{f}_{_{X_T, Y_T}}(\cdot)$ to the training splits $(X_T, Y_T)$ with the \gls{MO} $Y_T$ as the target. For each point in a validation split $(X_V, Y_V)$, its non-dominating rank $r$ in the split is calculated. The surrogates estimate the objective values on the validation split, $\hat{Y}_V=\hat{f}_{_{X_T,  Y_T}}(X_V)$, and their non-dominating ranks are calculated, producing predicted ranks $\hat{r}$. The performance of the predicted non-dominated ranks is evaluated using any error metric $\mathcal{E}(\hat{r}, r)$, that is constrained to sum to one. Finally, the non-dominated score can be calculated by aggregating the errors calculated for the different splits:
\gls{ParDen-Sur} allows the use of any metric $\mathcal{E}(\hat{r}, r)$ constrained to sum to one, to compare $r$ with $\hat{r}$:
\begin{equation}
\operatorname{NDScore}(\cdot) = \operatorname*{aggregation}_{\substack{(X_T, Y_T),(X_V, Y_V)\\ \in \operatorname{CV}_k(X)}} \mathcal{E}\left(\hat{r}, r\right) \\
\end{equation}
with
\begin{equation}
\hat{r} = \lceil Y_V \rceil^{\uparrow} \quad \mathrm{and} \quad
      r = \lceil\hat{f}_{_{X_T, Y_T}}(X_V) \rceil^{\uparrow}
\end{equation}
where $\lceil\cdot\rceil^\uparrow$ assigns the non-dominating rank to each vector. Since \gls{ParDen-Sur} uses the $k$-fold cross-validation function $\operatorname{CV}_k(\cdot)$ to estimate this value, the mean error across the $k$ splits is a suitable aggregation.

The presented method, \gls{ParDen-Sur}, recognises several deficiencies in the originally developed ParDen algorithm~\cite{van2021parden} and contains several extensions to overcome these. First, for the metric $\mathcal{E}\left(\hat{r}, r\right)$, \gls{ParDen-Sur} uses the Kendall-$\tau$ rank instead of accuracy. Second, \gls{ParDen-Sur} introduces a look-ahead mechanism that uses the surrogate model to iterate the \gls{EA} forward several generations. 

\gls{ParDen-Sur} is especially cognisant of the limitations of adding additional free parameters to a surrogate assisted algorithmic framework. To limit the number of free parameters and ensure diversity of the population, \gls{ParDen-Sur} uses reservoir sampling during the look-ahead procedure. Introducing a look-ahead procedure allows one to consider forgoing the acceptance sampling\footnote{Forgoing acceptance sampling without look-ahead is equivalent to the original \gls{EA} algorithm}. Algorithm~\ref{alg:pardensur} presents the full details of the proposed method with Figure~\ref{fig:parden_framework} giving a high-level overview of the details presented in the algorithm.

\begin{algorithm}[htbp]
    \caption{\gls{ParDen-Sur} Algorithm}\label{alg:pardensur}
    \SetKwInput{KwInput}{Input}                
    \SetKwInput{KwOutput}{Output}              
    \DontPrintSemicolon
    \SetNoFillComment

    \KwInput{\textit{\gls{EA}} ($M$), \textit{Terminating Criterion} ($\mathrm{TC}$), \textit{Look Ahead Terminating Criterion} ($\mathrm{TC}_L$), \textit{Simulation} ($f_e$), \textit{Surrogate} ($\mathcal{H}$), \textit{Loss} ($L$), \textit{Population Size} ($m$), \textit{Use Acceptance Sampling} (acceptance), \textit{Use Look Ahead} (look-ahead)}
    \KwOutput{\textit{Approximate Optimal Frontier ($\mathcal{P}$)}}
    \KwData{\textit{Daily Trading Data} ($D$)}

    \SetKwFunction{FMain}{main}
    \SetKwFunction{FLook}{look-ahead}
    \SetKwFunction{FMax}{fill}
    \SetKwFunction{FStep}{trainstep}
    \SetKwFunction{Init}{init}
    \SetKwFunction{FSub}{Sub}

    \SetKwProg{Fn}{Function}{:}{\KwRet}
    \Fn{\FLook{$\hat{f},\mathcal{G}, \ldots$}}{
        $X_P = \{\}$        
        \tcc*[r]{Initialise empty Reservoir}
        $r = m\times \operatorname{NDScore}(\hat{f}, \mathcal{G})$
        \tcc*[r]{Set Reservoir size as \% of non-dominated score}
        $M_L = \mathrm{copy}(M)$
        \tcc*[r]{Make throw away copy of \gls{EA}}
        $\mathcal{P}_C = \mathrm{copy}(G)$
        \tcc*[r]{Make throw away copy of Ground-truth}
        \While{$\mathrm{firstiteration}$ $\mathbf{or}$ $\neg \mathrm{TC}_L(M_L)$}{
            $X_L \sim M_L.\mathrm{infill}(m)$
            \tcc*[r]{Generate m look-ahead Candidates}
            $\mathcal{C}=\{(X_L,\hat{f}(X_L))\}$
            \tcc*[r]{Estimate Candidates' with Surrogate}
            $M_L.\operatorname{advance}(\mathcal{C})$
            \tcc*[r]{Advance \gls{EA} with Candidates}
            \tcc*[l]{Pretenders are non-dominated Candidates}
            $(X_C, \hat{Y}_C) = \mathcal{P}_C = \lceil \{\mathcal{C} \cup \mathcal{P}_C\} \rceil \cap C$

            $X_P = \operatorname{reservoir}(X_P, X_C, r)$
            \tcc*[r]{Reservoir sampling for Pretenders}
        }
        \KwRet $X_P$
    }



    \SetKwProg{Fn}{Main}{:}{\KwRet}
    \Fn{\FMain{}}{
    $\mathcal{G}=\{\}$
    \tcc*[r]{Init Ground-truth}
    $X_P \sim M.\mathrm{infill}(m)$
    \tcc*[r]{Generate m warm start pretenders}
    \While{$\mathrm{firstiteration}$ $\mathbf{or}$ $\neg \mathrm{TC}(M)$}{        
        $Y_P=f_e(X_P,D)$
        \tcc*[r]{Get pretenders' actual fitness}
        $\mathcal{G} = \lbrace\mathcal{G}\cup  \lbrace(X_P,Y_P)\rbrace\rbrace$
        \tcc*[r]{Add pretenders to ground-truth}
        $\hat{f} = \operatorname{argmin}_{h \in \mathcal{H}} \mathbb{E}[L(h,\mathcal{G})]$
        \tcc*[r]{Train surrogate on Ground-truth}    
        $M.\operatorname{advance}(\{(X_P, Y_P)\})$
            \tcc*[r]{Advance the \gls{EA}}    
            \If{$\mathrm{look-ahead}$}{               
                $X_P =\;$\FLook{$\hat{f}, \mathcal{G}$}
                \tcc*[r]{look-ahead for Pretenders}
            }\Else{
                $X_L \sim M_L.\mathrm{infill}(m)$
                \tcc*[r]{Generate m Candidates}
                $\mathcal{C}=\{(X_L,\hat{f}(X_L))\}$
                \tcc*[r]{Estimate Candidates' with Surrogate}                
                $(X_P, \hat{Y}_P) = \lceil \{\mathcal{C} \cup \mathcal{G}\} \rceil \cap C$
                \tcc*[r]{Pretenders are non-dominated Candidates}
            }
            $s = m - |X_P|$
            \tcc*[r]{Get remaining spaces s in Pretenders}
            
            \If{$\mathrm{acceptance}$}{
                \tcc*[l]{Acceptance sampling with non-dominated score threshold}
                $X_F = \operatorname{acceptance}(\sim M.\mathrm{infill}(s), \operatorname{NDScore}(\hat{f}, \mathcal{G}))$
            }\Else{
                $X_F \sim M.\mathrm{infill}(s)$
                \tcc*[r]{Generate s fill Candidates}                
            }
            $X_P = \{ X_P \cup X_F\}$
    }

        \KwRet $\mathcal{P} = \lceil \mathcal{G} \rceil$
    \tcc*[r]{Non-dominated solutions}
    }
\end{algorithm}

\tikzstyle{startstop} = [rectangle, rounded corners, minimum width=2cm, minimum height=1cm,text centered, draw=Set1-I, drop shadow, fill=Set1-I!50,]
\tikzstyle{io} = [trapezium, trapezium left angle=70, trapezium right angle=110, minimum width=2cm, minimum height=1cm, text centered, draw=black, drop shadow]
\tikzstyle{process} = [rectangle, minimum width=4.5cm, minimum height=1cm, text centered, text width=4cm, draw=Set1-E, drop shadow, fill = Set1-E!30]
\tikzstyle{process2} = [rectangle, minimum width=2cm, minimum height=1cm, text centered, text width=2cm, draw=black, drop shadow, fill = white]
\tikzstyle{process3} = [rectangle, minimum width=4.0cm, minimum height=1cm, text centered, text width=4.0cm, draw=Set1-E, drop shadow, fill = Set1-E!30]
\tikzstyle{process4} = [rectangle, minimum width=4.5cm, minimum height=1cm, text centered, text width=4.0cm, draw=Set1-B, drop shadow, fill = Set1-B!30]
\tikzstyle{decision} = [diamond, minimum width=3cm, minimum height=1cm, text centered, draw=Set1-A, drop shadow, fill = Set1-A!30]
\tikzstyle{arrow} = [thick,->,>=stealth]
\tikzset{Tank/.style={draw=black,fill=white,thick,rectangle,rounded corners=10pt,minimum width=0.75cm,minimum height=1.5cm,text width=0.75cm,align=center, drop shadow}}
\tikzstyle{surround} = [fill=gray!15,thick,draw=black,rounded corners=2mm]

\tikzstyle{line} = [draw, -stealth]

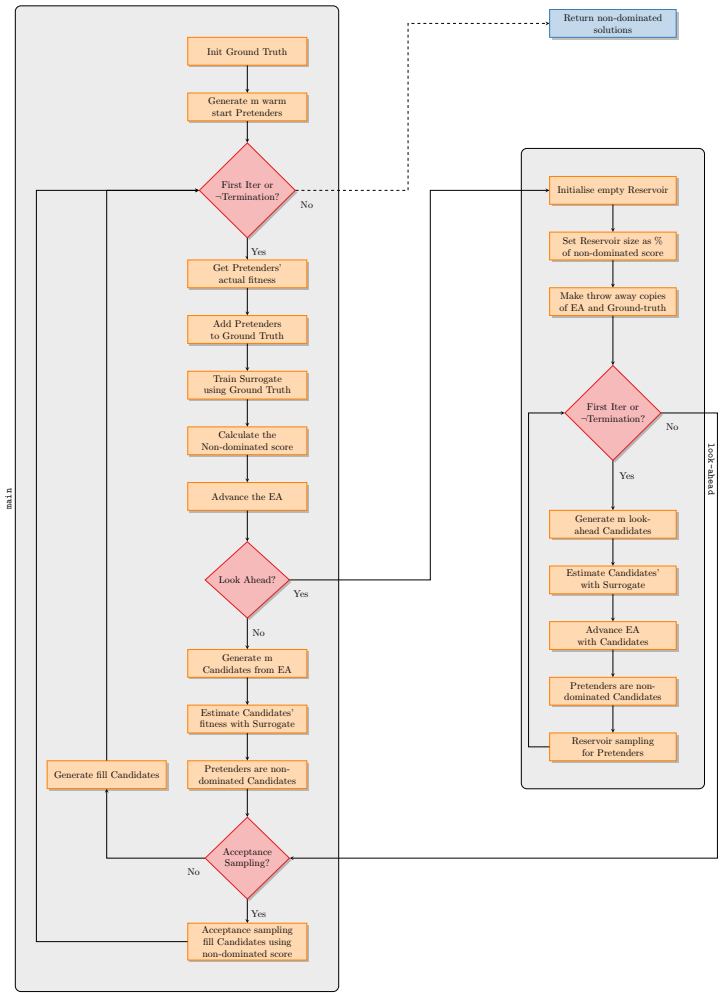
\begin{figure}[htbp]
    \centering
    \resizebox{.8\textwidth}{!}{
        \begin{tikzpicture}[node distance=1cm,every node/.style={fill=white}, align=center]


            \node (start) [xshift=3cm] {};


            \node (pro1) [process, below of = start, yshift=-2cm] {Initialise empty Reservoir};
            \node (pro2) [process, below of = pro1, yshift=-1cm] {Set Reservoir size as \% of non-dominated score};
            \node (pro3) [process, below of = pro2, yshift=-1cm] {Make throw away copies of \gls{EA} and Ground-truth};
            \node (proDec) [decision, below of = pro3, yshift=-3cm] {First Iter or\\$\neg$Termination?};
            \node (pro4) [process, below of = proDec, yshift=-3cm] {Generate m look-ahead Candidates};
            \node (pro5) [process, below of = pro4, yshift=-1cm] {Estimate Candidates' with Surrogate};
            \node (pro6) [process, below of = pro5, yshift=-1cm] {Advance \gls{EA} with Candidates};
            \node (pro7) [process, below of = pro6, yshift=-1cm] {Pretenders are non-dominated Candidates};
            \node (pro8) [process, below of = pro7, yshift=-1cm] {Reservoir sampling for Pretenders};

            \begin{pgfonlayer}{background}
                \node[surround] (background1) [inner sep=1.0cm, fit = (pro1) (pro8), label={[rotate=270, anchor=south]right:\texttt{look-ahead}}] {};
            \end{pgfonlayer}



            \node (main1) [process3, left of =pro1, xshift=-12cm, yshift=5cm] {Init Ground Truth};
            \node (main2) [process3, below of =main1, yshift=-1cm] {Generate m warm start Pretenders};
            \node (mainDec) [decision, below of=main2, yshift=-2.0cm] {First Iter or\\$\neg$Termination?};
            \node (main3) [process3, below of =mainDec, yshift=-2.0cm] {Get Pretenders' actual fitness};
            \node (main4) [process3, below of =main3, yshift=-1cm] {Add Pretenders to Ground Truth};
            \node (main5) [process3, below of =main4, yshift=-1cm] {Train Surrogate using Ground Truth};
            \node (main6) [process3, below of =main5, yshift=-1cm] {Calculate the Non-dominated score};
            \node (main7) [process3, below of =main6, yshift=-1cm] {Advance the \gls{EA}};
            \node (mainDec2) [decision, below of=main7, yshift=-2.0cm] {Look Ahead?};
            \node (main9) [process3, below of =mainDec2, yshift=-2.0cm] {Generate m Candidates from \gls{EA}};
            \node (main12) [process3, below of=main9, yshift=-1cm] {Estimate Candidates' fitness with Surrogate};
            \node (main13) [process3, below of =main12, yshift=-1.0cm] {Pretenders are non-dominated Candidates};
            \node (mainDec3) [decision, below of=main13, yshift=-2.0cm] {Acceptance\\Sampling?};
            \node (main14) [process3, left of =main13, xshift=-4cm] {Generate fill Candidates};
            \node (main15) [process3, below of =mainDec3, yshift=-2.0cm] {Acceptance sampling fill Candidates using non-dominated score};
            \begin{pgfonlayer}{background}
                \node[surround] (background2) [inner sep=1.125cm, fit = (main1) (main14) (main15) (main14), label={[rotate=90, anchor=south]left:\texttt{main}}] {};
            \end{pgfonlayer}


            \draw [arrow] (pro1) -- (pro2);
            \draw [arrow] (pro2) -- (pro3);
            \draw [arrow] (pro3) -- (proDec);
            \draw [arrow] (proDec.east) node[below, xshift=0.4cm, yshift=-0.25cm, fill=gray!15]{No} |- ++ (2.0cm, -0cm) |- (mainDec3);
            \draw [arrow] (proDec) node[below, xshift=0.5cm, yshift=-2.0cm, fill=gray!15]{Yes} -- (pro4);
            \draw [arrow] (pro4) -- (pro5);
            \draw [arrow] (pro5) -- (pro6);
            \draw [arrow] (pro6) -- (pro7);
            \draw [arrow] (pro7) -- (pro8);
            \draw [arrow] (pro8) -- ++ (-3cm, 0cm) |- (proDec.west);




            \draw [arrow] (main1) -- (main2);
            \draw [arrow] (main2) -- (mainDec);
            \draw [arrow] (mainDec.south) node[below, xshift=0.4cm, yshift=-0.25cm, fill=gray!15]{Yes} -- (main3);
            \draw [arrow] (main3) -- (main4);
            \draw [arrow] (main4) -- (main5);
            \draw [arrow] (main5) -- (main6);
            \draw [arrow] (main6) -- (main7);
            \draw [arrow] (main7) -- (mainDec2);
            \draw [arrow] (mainDec2.east) node[below, xshift=0.4cm, yshift=-0.25cm, fill=gray!15]{Yes} -- ++ (5cm, 0) |- (pro1.west);
            \draw [arrow] (mainDec2.south) node[below, xshift=0.4cm, yshift=-0.25cm, fill=gray!15]{No} -- (main9);
            \draw [arrow] (main9) -- (main12);
            \draw [arrow] (main12) -- (main13);
            \draw [arrow] (main13) -- (mainDec3);
            \draw [arrow] (mainDec3.west) node[below, xshift=-0.4cm, yshift=-0.25cm, fill=gray!15]{No} -| (main14);
            \draw [arrow] (mainDec3.south) node[below, xshift=0.4cm, yshift=-0.25cm, fill=gray!15]{Yes} -- (main15);
            \draw [arrow] (main14) |- (mainDec);
            \draw [arrow] (main15) -- ++(-7.5cm, 0) |- (mainDec);


            \node (out1) [process4, above of =start, yshift=2cm] {Return non-dominated solutions};
            \draw [arrow, dashed] (mainDec.east) node[below, xshift=0.4cm, yshift=-0.25cm, fill=gray!15]{No} |- ++ (4cm, 0cm) |- (out1);
        \end{tikzpicture}
    }
    \caption{Flowchart for the \gls{ParDen-Sur} Algorithm.}
    \label{fig:parden_framework}
\end{figure}

\section{Methodology}\label{sec:methodology}

In this study, the experimentation is split into two stages, referred to as the \textit{motivating} and \textit{primary} examples, respectively, for both the \gls{SPO} and \gls{MPO} cases. The motivating example is a scaled-down version of the problem presented in the primary example and is first explored to determine the most appropriate application of \gls{ParDen-Sur} within the primary example. In the motivating example, several well-known \gls{MOP} \glspl{EA} are initially applied to the aforementioned problem (see Section \ref{sec:prob_statement}) to determine which performs best and to provide a relevant baseline. After which, the study then applies \gls{ParDen-Sur} with the two best performing \glspl{EA} in the primary example.

All experiments are repeated ten times and average results are compared to the baseline \glspl{EA}. The outperformance of the Parden-Sur algorithm is assessed for statistical significance using one-sided Mann-Whitney U tests and adjusted for multiple testing using the Simes-Hochberg procedure. Hyper-parameters for the \glspl{EA} are manually adapted from van Zyl \textit{et al.}~\cite{van2021parden}.

\subsection{Method Metrics}\label{sec:metrics}

This study considers three metrics as performance indicators for the solution set generated by each algorithm. These include the two modified distance calculations, Generational Distance Plus (GD+) and Inverted Generational Distance Plus (IGD+)~\cite{ishibuchi2015modified}, as well as the Hypervolume (HV)~\cite{fonseca2006improved}. For GD+ and IGD+, an optimal target set needs to be specified, which is specified in the results below as the Pareto frontier from a random grid search with 6510 hyper-parameters. For HV only a reference point needs to be provided which is chosen to be $(40, 0)$ in the risk-return space.

It is often noted in the literature that, when looking into the solution quality of evolutionary approaches, no single metric can give a complete picture of which approach performs best in all circumstances. For this reason, evaluation across several different metrics is conducted. These metrics are defined by \cite{eiben2015} as follows:
\begin{itemize}
    \item \textbf{\gls{SR}:} Success is defined as a required quality criterion that needs to be achieved, i.e. a solution within 99\% or 95\% of the known optimal value. In this case, the results consider the HV of the random 6510 grid search as the optimal solution. The success rate is defined as the percentage of algorithm executions in which the success criterion is met.
    \item \textbf{\gls{AESR}:} Average number of evaluations of the \gls{EA} required to reach a solution at a given \gls{SR}, i.e. the \gls{AESR} at a \gls{SR} of 99\% would be the average number of evaluations it took to reach the stated \gls{SR}. The particular runs of the \gls{EA} which do not meet the \gls{SR} criterion are ignored.
    \item \textbf{\gls{AGSR}:} Average number of generations of the \gls{EA} required to reach a solution at a given \gls{SR}, i.e. the \gls{AGSR} at a \gls{SR} of 99\% would be the average number of generations it took to reach there. The particular runs of the \gls{EA} which do not meet the \gls{SR} criterion are ignored.
\end{itemize}

\subsection{Hyper-parameters}

The experiments use Latin Hyper-cube Sampling for random selection for all the \glspl{EA}, real Uniform Crossover (UX) for crossover, and real Polynomial Mutation (PM) for mutation. For all comparisons, the results use a population size of 60 and an offspring size of 30. The \glspl{EA} techniques all had a total evaluation budget of $510$. The study uses Riesz s-Energy to generate reference directions or reference points. Table~\ref{tab:hyper_parameters} shows the selected hyper-parameter values that are unique to the individual \glspl{EA}. In the study the hyper-parameters are selected once and these values are re-used for both the \gls{SPO} and \gls{MPO} problems~\cite{blank2020pymoo}.

\begin{table}[htbp]    
    \caption{Hyper-parameters used for the various \glspl{EA}.}\label{tab:hyper_parameters}
    \centering 
    \resizebox{.85\columnwidth}{!}{%
    \begin{tabular}{llcclr}      
         \toprule
         Dataset &   
         Method &
         \makecell[cc]{Crossover\\Rate} & 
         \makecell[cc]{Mutation\\Rate} & 
         Selection &
         Parameters \\         
         \midrule
         \multirow{6}{*}{DJIA} 
              & MO-CMA-ES  & -   & -   & -             & $\sigma = 0.1$ \\
              & NSGA-II    & 0.9 & 0.3 & Rank \& Crowd & - \\ 
              & NSGA-III   & 0.9 & 0.3 & Tournament(2) & - \\ 
              & R-NSGA-II  & 0.9 & 0.3 & -             & $\epsilon = 0.1$ \\ 
              & R-NSGA-III & 0.9 & 0.3 & Tournament(2) & $\mu = 0.1$ \\ 
              & U-NSGA-III & 0.9 & 0.3 & Tournament(2) & - \\          
         &&&&& \\
         \multirow{6}{*}{S\&P}
              & MO-CMA-ES  & -   & -   & -             & $\sigma = 0.1$ \\
              & NSGA-II    & 0.9 & 0.2 & Rank \& Crowd & - \\ 
              & NSGA-III   & 0.9 & 0.2 & Tournament(2) & - \\ 
              & R-NSGA-II  & 0.9 & 0.2 & -             & $\epsilon = 0.1$ \\ 
              & R-NSGA-III & 0.9 & 0.2 & Tournament(2) & $\mu = 0.1$ \\ 
              & U-NSGA-III & 0.9 & 0.2 & Tournament(2) & - \\ 
        \bottomrule
    \end{tabular}
    }
\end{table}

For \gls{ParDen-Sur}, the free parameter is the termination criteria for the look-ahead reservoir sampling. Multi-objective space tolerance termination was selected with a value of $0.0001$~\cite{blank2020pymoo} set according to the scale of the solutions which are values in the range 0 to 60 with up to three decimal places being significant.

\subsection{Research Instruments and Data}\label{sec:data_analysis}
The motivating and primary examples for both the SPO and MPO cases are prepared using two different datasets. The dataset used in the primary examples is prepared exactly as in Boyd \textit{et al.} \cite{boyd2017multi}, whereas a scaled-down version of this dataset is used in the motivating example. 

In summary, in the primary example, consideration is limited to assets from the S\&P 500 Index that has a full set of daily price data over the five years from January 2012 to December 2016. Daily volatilities used for determining transaction costs are calculated using the absolute difference between log open and close prices, while the bid-ask spread is fixed at $0.05\%$. Holding costs are fixed at $0.01\%$. These quantities are used to calculate the backtest performance of a policy.

Furthermore, risk, return, and trading cost forecasts are required at each period in the backtest period to construct the portfolio. Investigating different forecasting techniques is out of the scope of this study. Instead, return forecasts are imitated by adding zero-mean Gaussian noise to realised returns, whilst volatility and volume forecasts are specified using ten-day moving averages.

The datasets for the motivating example are prepared similarly, except that consideration is limited to assets from the Dow Jones Industrial Index over the eighteen months from July 2015 to December 2016. For further details, consult the supplementary materials.

All experiments use Ubuntu 20.04.1 LTS and are run on Intel(R) Xeon(R) CPU E5-2683 and E5-2695 v4 @ 2.10GHz. The study makes use of pymoo v0.5, pysamoo v0.1, and cvxportfolio \cite{boyd2017multi}. The source code and data for reproduction of the results are available on GitHub, \href{https://github.com/intelligent-systems-modelling/surrogate-assisted-moo-cvxport}{here}.

\section{Results}\label{sec:results}

 The following section discusses the results first for the \gls{SPO} and then for the \gls{MPO}.

\glsreset{SPO}
\subsection{\gls{SPO}}

\subsubsection{Motivating Example}\label{sec:spo_motivating}

One of the objectives of the motivating example is to ascertain which two of the various seminal \gls{MO} \gls{EA} optimisation techniques are most appropriate for applying the \gls{ParDen-Sur} algorithm within the \gls{SPO} problem. Considering Table~\ref{tab:spo_motivating_performance_indicators} we note that the results for NSGA-II and R-NSGA-II are mixed, with neither method outperforming the other across all indicators. It is worth noting that none of the implemented methods outperforms the 6510 random points.

Having identified the top-performing \gls{MO} \gls{EA} optimisation, the second objective of the motivating example is to determine which of the two selected methods is bested suited for the application of \gls{ParDen-Sur}. The results in Table~\ref{tab:spo_motivating_performance_indicators} highlight the superiority of ParDen-Sur applied to NSGA-II over R-NSGA-II. Further, the statistical tests indicate that \gls{ParDen-Sur} results are indeed significant and support an argument for \gls{ParDen-Sur}'s superiority over the baseline \gls{EA} as well as the superiority of using look-ahead with reservoir sampling over the original \gls{ParDen-Sur} algorithm. Further, the results show that no matter which underpinning \gls{EA} is used, the \gls{ParDen-Sur} results are always statistically significantly superior.

We consider the quality indicators next, where the objective is to ascertain the degree to which \gls{ParDen-Sur} is assisting the underpinning \gls{EA}. Table~\ref{tab:spo_motivating_quality_indicators} reinforces the selection of NSGA-II over R-NSGA-II given the superior \gls{SR} and similar total evaluations as measured by \gls{AESR} and \gls{AGSR}. Further, the results show that no matter which underpinning \gls{EA} is used, the superiority of \gls{ParDen-Sur} results are always statistically significant. 

\begin{table}[htbp]
    \begin{threeparttable}
        \caption{\gls{SPO} (Motivating Example): Performance indicators for the best observed value.}
        \label{tab:spo_motivating_performance_indicators}
        \centering
        \begin{tabular}{lrrr}
            \toprule
            Method                      & GD$+$          & IGD$+$         & HV             \\
            \midrule
            Random $6510$               & -              & -              & 925.8          \\
            Grid Search $510$~\cite{nystrup2020hyperparameter}           & .4488          & .4339          & 883.1          \\
            NSGA-II                     & .2298          & .2327          & 920.5 \\
            R-NSGA-II                   & .1272          & .4023          & 920.1          \\
            NSGA-III                    & .3646          & .4283          & 889.6          \\
            R-NSGA-III                  & .1987          & .3949          & 904.8          \\
            U-NSGA-III                  & .3699          & .3048          & 898.9          \\
            MO-CMA-ES                   & .3793          & .5156          & 882.4          \\
            &&&\\
            $\mathcal{P}$:NSGA-II$^-$   & .1442          & \rf{.1359}     & 933.8          \\
            $\mathcal{P}$:NSGA-II$^+$   & \bf{.0963}     & .1443          & \bf{938.3}     \\
            $\mathcal{S}$:NSGA-II$^+$   & .1329          & \bf{.1127}     & \rf{935.9}     \\
            $\mathcal{P}$:R-NSGA-II$^-$ & .1856          & .1512          & 929.5          \\
            $\mathcal{P}$:R-NSGA-II$^+$ & .1343          & .2385          & 922.8          \\
            $\mathcal{S}$:R-NSGA-II$^+$ & \rf{.1061}     & .1631          & 931.9          \\
            \bottomrule
        \end{tabular}
        \smallskip
        \begin{tablenotes}
            \item[$\mathcal{P}$] ParDen-Sur with acceptance sampling
            \item[$\mathcal{S}$] ParDen-Sur without acceptance sampling
            \item[$-$] Algorithm applied with no-look
            \item[$+$] Algorithm applied with look-ahead
        \end{tablenotes}
    \end{threeparttable}
\end{table}

\begin{table}[htbp]
    \begin{threeparttable}
        \caption{\gls{SPO} (Motivating Example): Quality Indicators. Bold: best, grey: top two, red: warning indicator.}
        \label{tab:spo_motivating_quality_indicators}
        \centering
        \begin{tabular}{lrrrrr}
            \toprule
            Method                      & SR$@95$ & SR$@90$   & AESR$@95$      & AESR$@90$ & AGSR$@95$    \\
            \midrule
            Grid Search $510$           & \rw{0.0} & \rw{0.0} & -              & -         & -            \\
            NSGA-II                     & 100.0    & 100.0    & 330.0          & 138.0     & 10.0         \\
            R-NSGA-II                   & \rw{90.0}& 100.0    & 283.4          & 144.0     & 8.5          \\
            NSGA-III                    & \rw{90.0}& 100.0    & 353.4          & 135.0     & 10.8         \\
            R-NSGA-III                  & \rw{60.0}& 100.0    & 293.4          & 131.0     & 8.7          \\
            U-NSGA-III                  & \rw{80.0}& 100.0    & 345.0          & 123.0     & 10.5         \\
            MO-CMA-ES                   & \rw{30.0}& \rw{60.0}& 410.0          & 203.4     & 12.7         \\
            &&&&&\\
            $\mathcal{P}$:NSGA-II$^-$   & 100.0     & 100.0   & 262.6          & 84.7      & 11.7     \\
            $\mathcal{P}$:NSGA-II$^+$   & \rw{80.0} & 100.0   & \rf{172.1}     & 80.1      & 10.3     \\
            $\mathcal{S}$:NSGA-II$^+$   & 100.0     & 100.0   & 177.0          & 78.0      & \bf{4.9} \\
            $\mathcal{P}$:R-NSGA-II$^-$ & 100.0     & 100.0   & 267.7          & 80.9      & 11.2     \\
            $\mathcal{P}$:R-NSGA-II$^+$ & \rw{40.0} & 100.0   & \bf{163.0}     & 114.5     & 10.0     \\
            $\mathcal{S}$:R-NSGA-II$^+$ & \rw{60.0} & 100.0   & 300.0          & 156.0     & \rf{9.0} \\
            \bottomrule
        \end{tabular}
        \smallskip
        \begin{tablenotes}
            \item[$\mathcal{P}$] ParDen-Sur with acceptance sampling
            \item[$\mathcal{S}$] ParDen-Sur without acceptance sampling
            \item[$-$] Algorithm applied with no-look
            \item[$+$] Algorithm applied with look-ahead
        \end{tablenotes}
    \end{threeparttable}
\end{table}

\subsubsection{Primary Example}\label{sec:spo_primary}

The main objective of the \gls{SPO} primary example is to replicate the existing research and provide context for the current \gls{ParDen-Sur} results. To this end Table~\ref{tab:spo_primary_performance_indicators} replicates the previous studies by \cite{nystrup2020hyperparameter} and \cite{van2021parden}. The results reinforce the superior performance of \gls{ParDen-Sur} over both NSGA-II and MO-CMA-ES as underpinning \glspl{EA}. Further, the results highlight the superiority of look-ahead with reservoir sampling as found in \gls{ParDen-Sur} over using just acceptance sampling. One interesting result requiring further investigation is the superior performance of R-NSGA-II when considering GD$+$.

Looking toward the quality indicators, the results in Table~\ref{tab:spo_primary_quality_indicators} reinforce the superior performance of \gls{ParDen-Sur} with look-ahead over \gls{ParDen-Sur} without for both NSGA-II and MO-CMA-ES. Having established the superiority of \gls{ParDen-Sur} with look-ahead in the \gls{SPO} problem space, the study now turns to the more challenging \gls{MPO} problem.

\begin{table}[htbp]
    \begin{threeparttable}
        \caption{\gls{SPO} (Primary Example): Performance indicators for the best observed value.}
        \label{tab:spo_primary_performance_indicators}
        \centering
        \begin{tabular}{lrrr}
            \toprule
            Method                                             & GD$+$      & IGD$+$     & HV          \\
            \midrule
            Random $6510$                                      & -          & -          & 1313.0      \\
            Grid Search $510$ & .5941      & .5703      & 1259.1      \\
            NSGA-II~\cite{van2021parden}                       & .1137      & .1585      & 1308.4      \\
            R-NSGA-II~\cite{van2021parden}                     & \bf{.0259} & .2443      & 1309.7      \\
            NSGA-III~\cite{van2021parden}                      & .1062      & .2320      & 1305.0      \\
            R-NSGA-III~\cite{van2021parden}                    & .1215      & .2279      & 1304.3      \\
            U-NSGA-III~\cite{van2021parden}                    & .1314      & .2567      & 1302.7      \\
            MO-CMA-ES~\cite{nystrup2020hyperparameter}         & .1002      & .0965      & 1317.4      \\
            &&&\\
            $\mathcal{P}$:MO-CMA-ES$^-$~\cite{van2021parden}   & .0912      & .0951      & 1320.0      \\
            $\mathcal{S}$:MO-CMA-ES$^+$                        & .0602      & \rf{.0757} & \rf{1323.5} \\
            $\mathcal{P}$:NSGA-II$^-$~\cite{van2021parden}     & .0764      & .0901      & 1319.4      \\
            $\mathcal{S}$:NSGA-II$^+$                          & \rf{.0558} & \bf{.0604} & \bf{1323.9} \\
            \bottomrule
        \end{tabular}
        \smallskip
        \begin{tablenotes}
            \item[$\mathcal{P}$] ParDen-Sur with acceptance sampling
            \item[$\mathcal{S}$] ParDen-Sur without acceptance sampling
            \item[$-$] Algorithm applied with no-look
            \item[$+$] Algorithm applied with look-ahead
        \end{tablenotes}
    \end{threeparttable}
\end{table}

\begin{table}[htbp]
    \begin{threeparttable}
        \caption{\gls{SPO} (Primary Example): Quality Indicators. Bold: best, grey: top two, red: warning indicator.}
        \label{tab:spo_primary_quality_indicators}
        \centering
        \begin{tabular}{lrrrrr}
            \toprule
            Method                                             & SR$@99$ & SR$@95$ & AESR$@99$  & AESR$@95$ & AGSR$@99$ \\
            \midrule
            Grid Search $510$~\cite{nystrup2020hyperparameter} & \rw{0.0} & 100.0   & -          & 510.0     & -         \\
            NSGA-II~\cite{van2021parden}                       & 100.0    & 100.0   & 327.0      & 63.0      & 9.9       \\
            R-NSGA-II~\cite{van2021parden}                     & 100.0    & 100.0   & 297.0      & 63.0      & 8.9       \\
            NSGA-III~\cite{van2021parden}                      & 100.0    & 100.0   & 405.0      & 81.0      & 12.5      \\
            R-NSGA-III~\cite{van2021parden}                    & \rw{60.0}& 100.0   & 415.3      & 75.2      & 12.8      \\
            U-NSGA-III~\cite{van2021parden}                    & \rw{70.0}& 100.0   & 338.6      & 96.0      & 10.3      \\
            MO-CMA-ES~\cite{nystrup2020hyperparameter}         & 100.0    & 100.0   & 333.0      & 99.0      & 10.1      \\
            &&&&&\\
            $\mathcal{P}$:MO-CMA-ES$^-$~\cite{van2021parden}   & 100.0   & 100.0   & 247.7      & 75.9      & 10.7      \\
            $\mathcal{S}$:MO-CMA-ES$^+$                        & 100.0   & 100.0   & \rf{216.2} & 72.1      & \rf{6.8}  \\
            $\mathcal{P}$:NSGA-II$^-$~\cite{van2021parden}     & 100.0   & 100.0   & 254.7      & 66.4      & 10.1      \\
            $\mathcal{S}$:NSGA-II$^+$                          & 100.0   & 100.0   & \bf{177.0} & 60.0      & \bf{4.9}  \\
            \bottomrule
        \end{tabular}
        \smallskip
        \begin{tablenotes}
            \item[$\mathcal{P}$] ParDen-Sur with acceptance sampling
            \item[$\mathcal{S}$] ParDen-Sur without acceptance sampling
            \item[$-$] Algorithm applied with no-look
            \item[$+$] Algorithm applied with look-ahead
        \end{tablenotes}
    \end{threeparttable}
\end{table}

\glsreset{MPO}
\subsection{\gls{MPO}}

\subsubsection{Motivating Example}\label{sec:mpo_motivating}

The main objective of the motivating example was to establish which \gls{MO} \gls{EA} algorithms to use in conjunction with \gls{ParDen-Sur}. Table~\ref{tab:mpo_motivating_performance_indicators}, like in the \gls{SPO} motivating example, indicates that NSGA-II and R-NSGA-II are superior choices for underpinning \glspl{EA}. Both of these techniques significantly outperform MO-CMA-ES previously proposed by Nystrup \textit{et al.}~\cite{nystrup2020hyperparameter}.

When considering the second objective of the motivating example as to which of the top two underpinning \glspl{EA} are best suited to further exploration with \gls{ParDen-Sur}, Table~\ref{tab:mpo_motivating_performance_indicators} shows a slight benefit in favour of NSGA-II. However, when we consider the quality indicators in Table~\ref{tab:mpo_motivating_quality_indicators}, the results show that any gains by R-NSGA-II are also linked to substandard Success Rates at the $99\%$ threshold. Interestingly, the quality indicators indicate that \gls{ParDen-Sur} has lower \gls{AESR}@99 and \gls{AGSR}@99 than the underpinning \glspl{EA}. However, the jump in \gls{SR}@99 to 100\% warrants further exploration and points to the superiority of \gls{ParDen-Sur} not just for reducing the rate of convergence but also for improving the probability of convergence. This argument is further supported when considering the superiority of \gls{ParDen-Sur} at the 95\% threshold. 

Having established the superiority of \gls{ParDen-Sur} coupled with NSGA-II on the \gls{MPO} motivating example, the study now focuses on the \gls{MPO} primary results.

\begin{table}[htbp]
    \begin{threeparttable}
        \caption{\gls{MPO} (Motivating Example): Performance indicators for the best observed value.}
        \label{tab:mpo_motivating_performance_indicators}
        \centering
        \begin{tabular}{lrrr}
            \toprule
            Method                                     & GD$+$ & IGD$+$ & HV     \\
            \midrule
            Random $6510$                              & -     & -      & 1632.9 \\
            Grid Search $510$                          & .5497 & .8091  & 1530.1 \\
            NSGA-II                                    & .1261 & .1938  & 1630.5 \\
            R-NSGA-II                                  & .1206 & .2456  & 1615.8 \\
            NSGA-III                                   & .1412 & .2178  & 1619.9 \\
            R-NSGA-III                                 & .1775 & .3260  & 1607.7 \\
            U-NSGA-III                                 & .2213 & .2836  & 1600.3 \\
            MO-CMA-ES~\cite{nystrup2020hyperparameter} & .2360 & .6529  & 1555.4 \\
            &&&\\
            $\mathcal{P}$:NSGA-II$^-$                  & .0905      & .1266      & \bf{1643.6} \\
            $\mathcal{P}$:NSGA-II$^+$                  & .0910      & \bf{.1048} & \rf{1640.7} \\
            $\mathcal{S}$:NSGA-II$^+$                  & .1007      & .1306      & 1639.9 \\
            $\mathcal{P}$:R-NSGA-II$^-$                & .1083      & .1395      & 1635.9 \\
            $\mathcal{P}$:R-NSGA-II$^+$                & \bf{.0765} & \rf{.1159} & 1640.3 \\
            $\mathcal{S}$:R-NSGA-II$^+$                & \rf{.0845} & .1291      & 1635.4 \\
            \bottomrule
        \end{tabular}
        \smallskip
        \begin{tablenotes}
            \item[$\mathcal{P}$] ParDen-Sur with acceptance sampling
            \item[$\mathcal{S}$] ParDen-Sur without acceptance sampling
            \item[$-$] Algorithm applied with no-look
            \item[$+$] Algorithm applied with look-ahead
        \end{tablenotes}
    \end{threeparttable}
\end{table}

\begin{table}[htbp]
    \begin{threeparttable}
        \caption{\gls{MPO} (Motivating Example): Quality Indicators. Bold: best, grey: top two, red: warning indicator.}
        \label{tab:mpo_motivating_quality_indicators}
        \centering
        \begin{tabular}{lrrrrr}
            \toprule
            Method                      & SR$@99$ & SR$@95$ & AESR$@99$ & AESR$@95$ & AGSR$@99$ \\
            \midrule
            Grid Search $510$           & \rw{ 0.0} & 100.0    & -         & 510.0     & -         \\
            NSGA-II                     & \rw{80.0} & 100.0    & 315.0     & 138.0     & 9.5       \\
            R-NSGA-II                   & \rw{60.0} & 100.0    & 390.0     & 147.0     & 12.0      \\
            NSGA-III                    & \rw{10.0} & \rw{90.0}& 450.0     & 230.0     & 14.0      \\
            R-NSGA-III                  & \rw{10.0} & 100.0    & 452.0     & 191.0     & 14.0      \\
            U-NSGA-III                  & \rw{10.0} & 100.0    & 510.0     & 243.0     & 16.0      \\
            MO-CMA-ES                   & \rw{ 0.0} & \rw{40.0}& -         & 316.7     & -         \\
            &&&&&\\
            $\mathcal{P}$:NSGA-II$^-$   & 100.0     & 100.0   & 321.0     & 117.0     & 9.7       \\
            $\mathcal{P}$:NSGA-II$^+$   & 100.0     & 100.0   & 327.0     & 105.0     & 9.9       \\
            $\mathcal{S}$:NSGA-II$^+$   & 100.0     & 100.0   & 327.0     & 120.0     & 9.9       \\
            $\mathcal{P}$:R-NSGA-II$^-$ & 100.0     & 100.0   & 333.0     & 108.0     & 10.1      \\
            $\mathcal{P}$:R-NSGA-II$^+$ & \rw{90.0} & 100.0   & \rf{293.3}& 114.0     & \rf{8.8}  \\
            $\mathcal{S}$:R-NSGA-II$^+$ & \rw{90.0} & 100.0   & \bf{273.3}& 93.0      & \bf{8.1}  \\
            \bottomrule
        \end{tabular}
        \smallskip
        \begin{tablenotes}
            \item[$\mathcal{P}$] ParDen-Sur with acceptance sampling
            \item[$\mathcal{S}$] ParDen-Sur without acceptance sampling
            \item[$-$] Algorithm applied with no-look
            \item[$+$] Algorithm applied with look-ahead
        \end{tablenotes}
    \end{threeparttable}
\end{table}

\subsubsection{Primary Example}\label{sec:mpo_primary}

The main aim of this study is to evaluate the use of \gls{ParDen-Sur} in the context of \gls{MPO} on the primary example. Table~\ref{tab:mpo_primary_performance_indicators} shows the out-performance of all variants of \gls{ParDen-Sur}. Most interesting is the superior performance obtained using look-ahead. The p-values of the Mann-Whitney U tests (with multiple testing adjustments) in Table~\ref{tab:mpo_primary_stats} add further weight by demonstrating the out-performance is statistically significant.

\begin{table}[htb!]
    \centering
    \caption{MPO (Primary Example): $p$-values from testing differences in performance indicators between Parden-Sur and baseline methods.}
    \label{tab:mpo_primary_stats}
    \resizebox{\columnwidth}{!}{%
    \begin{tabular}{lrrrrrrrrr}
    \toprule
    {} & \multicolumn{3}{c}{HV} & \multicolumn{3}{c}{GD+} & \multicolumn{3}{c}{IGD+} \\
    \cmidrule{2-4}
    \cmidrule{5-7}
    \cmidrule{8-10}
    {} & $\mathcal{P}$:NSGA-II- & $\mathcal{P}$:NSGA-II+ & $\mathcal{S}$:NSGA-II & $\mathcal{S}$:NSGA-II- & $\mathcal{S}$:NSGA-II+ & $\mathcal{S}$:NSGA-II & $\mathcal{P}$:NSGA-II- & $\mathcal{P}$:NSGA-II+ & $\mathcal{S}$:NSGA-II \\
    \midrule
    NSGA-II    &   \num[round-precision=2,round-mode=figures, scientific-notation=true]{0.002040} &   \num[round-precision=2,round-mode=figures, scientific-notation=true]{0.000913} &  \num[round-precision=2,round-mode=figures, scientific-notation=true]{0.000913} &   \num[round-precision=2,round-mode=figures, scientific-notation=true]{0.021134} &   \num[round-precision=2,round-mode=figures, scientific-notation=true]{0.000913} &  \num[round-precision=2,round-mode=figures, scientific-notation=true]{0.000913} &   \num[round-precision=2,round-mode=figures, scientific-notation=true]{0.000913} &   \num[round-precision=2,round-mode=figures, scientific-notation=true]{0.000913} &  \num[round-precision=2,round-mode=figures, scientific-notation=true]{0.000913} \\
    R-NSGA-II  &   \num[round-precision=2,round-mode=figures, scientific-notation=true]{0.000913} &   \num[round-precision=2,round-mode=figures, scientific-notation=true]{0.000913} &  \num[round-precision=2,round-mode=figures, scientific-notation=true]{0.000913} &   \num[round-precision=2,round-mode=figures, scientific-notation=true]{0.037831} &   \num[round-precision=2,round-mode=figures, scientific-notation=true]{0.003287} &  \num[round-precision=2,round-mode=figures, scientific-notation=true]{0.010927} &   \num[round-precision=2,round-mode=figures, scientific-notation=true]{0.000913} &   \num[round-precision=2,round-mode=figures, scientific-notation=true]{0.000913} &  \num[round-precision=2,round-mode=figures, scientific-notation=true]{0.000913} \\
    NSGA-III   &   \num[round-precision=2,round-mode=figures, scientific-notation=true]{0.000913} &   \num[round-precision=2,round-mode=figures, scientific-notation=true]{0.000913} &  \num[round-precision=2,round-mode=figures, scientific-notation=true]{0.000913} &   \num[round-precision=2,round-mode=figures, scientific-notation=true]{0.007221} &   \num[round-precision=2,round-mode=figures, scientific-notation=true]{0.000913} &  \num[round-precision=2,round-mode=figures, scientific-notation=true]{0.001108} &   \num[round-precision=2,round-mode=figures, scientific-notation=true]{0.000913} &   \num[round-precision=2,round-mode=figures, scientific-notation=true]{0.000913} &  \num[round-precision=2,round-mode=figures, scientific-notation=true]{0.000913} \\
    R-NSGA-III &   \num[round-precision=2,round-mode=figures, scientific-notation=true]{0.000913} &   \num[round-precision=2,round-mode=figures, scientific-notation=true]{0.000913} &  \num[round-precision=2,round-mode=figures, scientific-notation=true]{0.000913} &   \num[round-precision=2,round-mode=figures, scientific-notation=true]{0.002040} &   \num[round-precision=2,round-mode=figures, scientific-notation=true]{0.000913} &  \num[round-precision=2,round-mode=figures, scientific-notation=true]{0.000913} &   \num[round-precision=2,round-mode=figures, scientific-notation=true]{0.000913} &   \num[round-precision=2,round-mode=figures, scientific-notation=true]{0.000913} &  \num[round-precision=2,round-mode=figures, scientific-notation=true]{0.000913} \\
    U-NSGA-III &   \num[round-precision=2,round-mode=figures, scientific-notation=true]{0.000913} &   \num[round-precision=2,round-mode=figures, scientific-notation=true]{0.000913} &  \num[round-precision=2,round-mode=figures, scientific-notation=true]{0.000913} &   \num[round-precision=2,round-mode=figures, scientific-notation=true]{0.002306} &   \num[round-precision=2,round-mode=figures, scientific-notation=true]{0.000913} &  \num[round-precision=2,round-mode=figures, scientific-notation=true]{0.000913} &   \num[round-precision=2,round-mode=figures, scientific-notation=true]{0.000913} &   \num[round-precision=2,round-mode=figures, scientific-notation=true]{0.000913} &  \num[round-precision=2,round-mode=figures, scientific-notation=true]{0.000913} \\
    MO-CMA-ES  &   \num[round-precision=2,round-mode=figures, scientific-notation=true]{0.000913} &   \num[round-precision=2,round-mode=figures, scientific-notation=true]{0.000913} &  \num[round-precision=2,round-mode=figures, scientific-notation=true]{0.000913} &   \num[round-precision=2,round-mode=figures, scientific-notation=true]{0.000913} &   \num[round-precision=2,round-mode=figures, scientific-notation=true]{0.000913} &  \num[round-precision=2,round-mode=figures, scientific-notation=true]{0.000913} &   \num[round-precision=2,round-mode=figures, scientific-notation=true]{0.000913} &   \num[round-precision=2,round-mode=figures, scientific-notation=true]{0.000913} &  \num[round-precision=2,round-mode=figures, scientific-notation=true]{0.000913} \\
    \bottomrule
    \end{tabular}}
\end{table}

When considering the quality indicators in Table~\ref{tab:mpo_primary_quality_indicators} the results show marked improvements in both \gls{AESR}@99 and \gls{AGSR}@99, with \gls{ParDen-Sur} having an almost 2$\times$ speedup over the underpinning \gls{EA} in both these metrics. We also note that \gls{ParDen-Sur} without acceptance sampling outperforms when considering the \gls{AESR}@99. This result is interesting as it suggests the benefit of one less free parameter and a possible algorithm simplification. It is worth contrasting this with the results in Table~\ref{tab:spo_motivating_quality_indicators} and \ref{tab:mpo_motivating_quality_indicators} which have both supporting and conflicting support of this observation.

\begin{table}[htbp]
    \begin{threeparttable}
        \caption{\gls{MPO} (Primary Example): Performance indicators for the best observed value.}
        \label{tab:mpo_primary_performance_indicators}
        \centering
        \begin{tabular}{lrrr}
            \toprule
            Method                                     & GD$+$          & IGD$+$         & HV              \\
            \midrule
            Random $6510$                              & -              & -              & 1699.9          \\
            Grid Search~\cite{boyd2017multi}           & .8852          & 1.3881         & 1487.7          \\
            Grid Search $510$                          & .4634          & .4534          & 1644.7          \\
            NSGA-II                                    & .0998          & .2110          & 1700.8          \\
            R-NSGA-II                                  & .0531          & .2168          & 1694.9          \\
            NSGA-III                                   & .0852          & .2361          & 1692.5          \\
            R-NSGA-III                                 & .1054          & .2329          & 1690.1          \\
            U-NSGA-III                                 & .1189          & .2378          & 1689.7          \\
            MO-CMA-ES~\cite{nystrup2020hyperparameter} & .1751          & .2248          & 1695.9          \\
            &&&\\
            $\mathcal{P}$:NSGA-II$^-$                  & .0602          & .0838          & 1711.2          \\
            $\mathcal{P}$:NSGA-II$^+$                  & \bf{.0428}     & \rf{.0731}     & \bf{1715.6}     \\
            $\mathcal{S}$:NSGA-II$^+$                  & \rf{.0490}     & \bf{.0706}     & \rf{1715.4}     \\
            \bottomrule
        \end{tabular}
        \smallskip
        \begin{tablenotes}
            \item[$\mathcal{P}$] ParDen-Sur with acceptance sampling
            \item[$\mathcal{S}$] ParDen-Sur without acceptance sampling
            \item[$-$] Algorithm applied with no-look
            \item[$+$] Algorithm applied with look-ahead
        \end{tablenotes}
    \end{threeparttable}
\end{table}

\begin{table}[htbp]
    \begin{threeparttable}
        \caption{\gls{MPO} (Primary Example): Quality Indicators. Bold: best, grey: top two, red: warning indicator.}
        \label{tab:mpo_primary_quality_indicators}
        \centering
        \begin{tabular}{lrrrrr}
            \toprule
            Method                    & SR$@99$ & SR$@95$ & AESR$@99$      & AESR$@95$     & AGSR$@99$    \\
            \midrule
            Grid Search~\cite{boyd2017multi}
                                      & \rw{0.0} & \rw{0.0}& -              & -             & -            \\
            Grid Search $510$         & \rw{0.0} & 100.0   & -              & 510.0         & -            \\
            NSGA-II                   & 100.0    & 100.0   & 309.0          & 72.0          & 9.3          \\
            R-NSGA-II                 & 100.0    & 100.0   & 321.0          & 72.0          & 9.7          \\
            NSGA-III                  & 100.0    & 100.0   & 354.0          & 75.0          & 10.8         \\
            R-NSGA-III                & \rw{70.0}& 100.0   & 404.9          & 92.0          & 12.4         \\
            U-NSGA-III                & \rw{90.0}& 100.0   & 383.3          & 81.0          & 11.8         \\
            MO-CMA-ES                 & \rw{20.0}& 100.0   & 435.0          & 93.0          & 13.5         \\
            &&&&&\\
            $\mathcal{P}$:NSGA-II$^-$ & 100.0   & 100.0   & 258.0          & 66.0          & 7.6          \\
            $\mathcal{P}$:NSGA-II$^+$ & 100.0   & 100.0   & \rf{171.0}     & 78.0          & \rf{4.7}     \\
            $\mathcal{S}$:NSGA-II$^+$ & 100.0   & 100.0   & \bf{168.0}     & 63.0          & \bf{4.6}     \\
            \bottomrule
        \end{tabular}
        \smallskip
        \begin{tablenotes}
            \item[$\mathcal{P}$] ParDen-Sur with acceptance sampling
            \item[$\mathcal{S}$] ParDen-Sur without acceptance sampling
            \item[$-$] Algorithm applied with no-look
            \item[$+$] Algorithm applied with look-ahead
        \end{tablenotes}
    \end{threeparttable}
\end{table}

\pgfplotstableread[col sep = comma]{mpo_mpo_nsga_ii_HV_plot_data.txt}\MPOnsgaii
\pgfplotstableread[col sep = comma]{mpo_pardensur_1.0_lookahead_mpo_nsga_ii_HV_plot_data.txt}\MPOpardenOnelookahead
\pgfplotstableread[col sep = comma]{mpo_pardensur_1.0_nolook_mpo_nsga_ii_HV_plot_data.txt}\MPOpardenOnenolook
\pgfplotstableread[col sep = comma]{mpo_pardensur_2.0_lookahead_mpo_nsga_ii_HV_plot_data.txt}\MPOpardenTwolookahead

\section{Discussion}\label{sec:discussion}

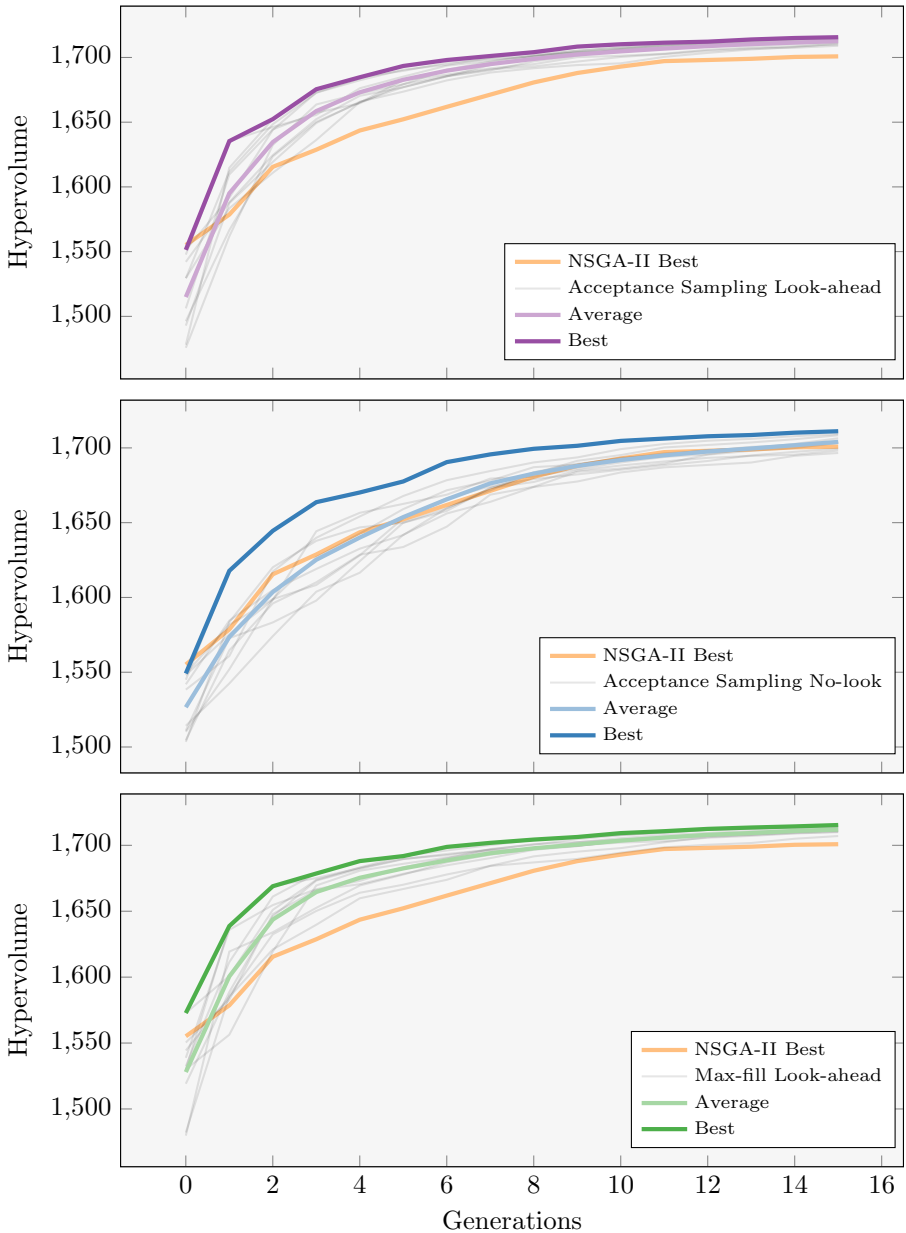
\begin{figure}[htb!]
    \centering
    \begin{subfigure}[htbp]{\textwidth}
    \begin{tikzpicture}
      \begin{axis}[
                xticklabels={},
                ylabel=Hypervolume,
                axis background/.style={fill=gray!7, draw=gray},
                width=\textwidth,
                height=.9*\axisdefaultheight,
                legend cell align={left},
                legend style={at={(0.99,0.05),font=\footnotesize},
                anchor=south east,legend columns=1}
        ]
        \addplot+[Set1-E!50, ultra thick] table[x expr=\coordindex, y index = {11}]{\MPOnsgaii};
        \addlegendentry{NSGA-II Best}
        
        \foreach \x in {0,...,8}
            \addplot[thick, opacity=0.1, forget plot]
            table[x expr=\coordindex, y index = {\x}]{\MPOpardenOnelookahead};
        \addplot[thick, opacity=0.1] table[x expr=\coordindex, y index = {9}]{\MPOpardenOnelookahead};
        \addlegendentry{Acceptance Sampling Look-ahead}

        \addplot+[Set1-D!50, ultra thick] table[x expr=\coordindex, y index = {10}]{\MPOpardenOnelookahead};
        \addlegendentry{Average}
        
        \addplot+[Set1-D, ultra thick] table[x expr=\coordindex, y index = {11}]{\MPOpardenOnelookahead};
        \addlegendentry{Best}
        
      \end{axis}
\end{tikzpicture}
\end{subfigure}%
\hfill
    \begin{subfigure}[htbp]{\textwidth}
    \begin{tikzpicture}
      \begin{axis}[
                xticklabels={},
                ylabel=Hypervolume,
                axis background/.style={fill=gray!7, draw=gray},
                width=\textwidth,
                height=.9*\axisdefaultheight,
                legend cell align={left},
                legend style={at={(0.99,0.05),font=\footnotesize},
                anchor=south east,legend columns=1}
        ]
        \addplot+[Set1-E!50, ultra thick] table[x expr=\coordindex, y index = {11}]{\MPOnsgaii};
        \addlegendentry{NSGA-II Best}
        
        \foreach \x in {0,...,8}
            \addplot[thick, opacity=0.1, forget plot]
            table[x expr=\coordindex, y index = {\x}]{\MPOpardenOnenolook};
        \addplot[thick, opacity=0.1] table[x expr=\coordindex, y index = {9}]{\MPOpardenOnenolook};
        \addlegendentry{Acceptance Sampling No-look}
        
        \addplot+[Set1-B!50, ultra thick] table[x expr=\coordindex, y index = {10}]{\MPOpardenOnenolook};
        \addlegendentry{Average}
        
        \addplot+[Set1-B, ultra thick] table[x expr=\coordindex, y index = {11}]{\MPOpardenOnenolook};
        \addlegendentry{Best}

      \end{axis}
\end{tikzpicture}
\end{subfigure}%
\hfill
    \begin{subfigure}[htbp]{\textwidth}
    \begin{tikzpicture}
      \begin{axis}[
                xlabel=Generations,
                ylabel=Hypervolume,
                axis background/.style={fill=gray!7, draw=gray},
                width=\textwidth,
                height=.9*\axisdefaultheight,
                legend cell align={left},
                legend style={at={(0.99,0.05),font=\footnotesize},
                anchor=south east,legend columns=1}
        ]
        \addplot+[Set1-E!50, ultra thick] table[x expr=\coordindex, y index = {11}]{\MPOnsgaii};
        \addlegendentry{NSGA-II Best}

        \foreach \x in {0,...,8}
            \addplot[thick, opacity=0.1, forget plot]
            table[x expr=\coordindex, y index = {\x}]{\MPOpardenTwolookahead};
        \addplot[thick, opacity=0.1] table[x expr=\coordindex, y index = {9}]{\MPOpardenTwolookahead};
        \addlegendentry{Max-fill Look-ahead}
        
        \addplot+[Set1-C!50, ultra thick] table[x expr=\coordindex, y index = {10}]{\MPOpardenTwolookahead};
        \addlegendentry{Average}
        
        \addplot+[Set1-C, ultra thick] table[x expr=\coordindex, y index = {11}]{\MPOpardenTwolookahead};
        \addlegendentry{Best}

      \end{axis}
\end{tikzpicture}
\end{subfigure}%
    \caption{\gls{ParDen-Sur} Variants vs NSGA-II Hypervolume.  All \gls{ParDen-Sur} variants with Look-ahead on average can obtain an optimal solution in fewer generations. Indicated by their positions in the upper left-hand corners. Trial lines are in grey.}\label{fig:mpo_parden_all}
\end{figure}

\begin{figure}[htb!]
    \centering
    \begin{subfigure}[htbp]{\textwidth}
    \begin{tikzpicture}
      \begin{axis}[
                xlabel=Generations,
                ylabel=Hypervolume,
                axis background/.style={fill=gray!7, draw=gray},
                width=\textwidth,
                height=.9*\axisdefaultheight,
                legend cell align={left},
                legend style={at={(0.99,0.05),font=\footnotesize},
                anchor=south east,legend columns=1}
        ]
        \addplot+[Set1-E!50, ultra thick] table[x expr=\coordindex, y index = {11}]{\MPOnsgaii};
        \addlegendentry{NSGA-II Best}

        \addplot+[Set1-D, ultra thick] table[x expr=\coordindex, y index = {11}]{\MPOpardenOnelookahead};
        \addlegendentry{Acceptance Sampling Look-ahead Best}
        
        \addplot+[Set1-B, ultra thick] table[x expr=\coordindex, y index = {11}]{\MPOpardenOnenolook};
        \addlegendentry{Acceptance Sampling No-look Best}
        
        \addplot+[Set1-C, ultra thick] table[x expr=\coordindex, y index = {11}]{\MPOpardenTwolookahead};
        \addlegendentry{Max-fill Look-ahead Best}

      \end{axis}
\end{tikzpicture}
\end{subfigure}%
    \caption{Best \gls{ParDen-Sur} Variants vs NSGA-II Hypervolume. The max-fill with Look-ahead can obtain an optimal solution in the fewest generations. Indicated by its dominant position in the upper left-hand corner.}\label{fig:mpo_parden_best}
\end{figure}
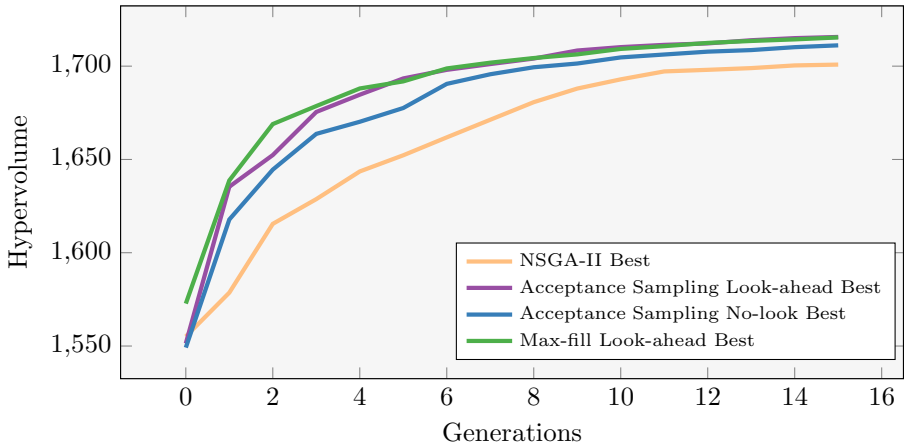

The presented results demonstrate that the challenges of optimal hyper-parameter search related to multi-period optimisation are tractable within the context of surrogate assisted \gls{MO} optimisation. Further, the results support the use of the \gls{ParDen-Sur} algorithm with NSGA-II to significantly improve the rate at which one can arrive at a solution and the quality of the solutions found. 

Previously, it has been shown that surrogate-assisted portfolio optimisation can speed up single-period hyper-parameter searches~\cite{van2021parden}. This study confirms those results and extends them to the multi-period optimisation use case. The outcome suggests that all versions of \gls{ParDen-Sur}, the proposed method, lead to a speed-up in acquiring optimal frontiers in the portfolio optimisation task, as seen in Figure~\ref{fig:mpo_parden_all}.

Previous work~\cite{van2021parden} showed that the ParDen framework was able to speed up both MO-CMA-ES and NSGA-II, and the results in this study show that this extends to R-NSGA-II. Further, the results are demonstrated on more than one dataset and for multiple periods. However, without testing this statement across the full battery of \glspl{EA}, it is not possible to ascertain if the extensions to ParDen as contained in \gls{ParDen-Sur} are a general purpose surrogate assisted optimisation framework.

In the previous study, ParDen used only acceptance sampling to improve speed-up~\cite{van2021parden}. Here, the results demonstrate that the inclusion of look-ahead with acceptance sampling is a superior approach, as confirmed by Figure~\ref{fig:mpo_parden_best}. Further, when \gls{AESR} is not concerned, forgoing acceptance sampling in favour of looking ahead provides the best speed-up.

It is worth noting that the combination of acceptance sampling and look-ahead has a net positive impact on \gls{AESR}. The case for minimal \gls{AESR} would occur when one is, for instance, being charged for resources. Here using the smallest amount of evaluations of the simulation is desirable over the smallest number of generations as measured by \gls{AGSR}.

\section{Conclusion}\label{sec:conclusion}

This study presents the \gls{ParDen-Sur} surrogate assisted optimisation algorithm, an extension to the original ParDen algorithm by van Zyl \textit{et al.}~\cite{van2021parden}, to address the computational difficulty of finding the efficient frontier for multi-period portfolio optimisation. The extension takes the form of a reservoir sampling-enabled look-ahead mechanism for generating new offspring. Both the look-ahead and the acceptance sampling rely on the NDScore. Here the NDScore is extended to include Kendall-$\tau$. The extension and the original acceptance sampling mechanism are tested on the problems of \gls{SPO} and \gls{MPO}. The three variants of \gls{ParDen-Sur} are evaluated across several \glspl{EA}. Both performance and quality metrics are taken into account.
The results demonstrate a statistically significant improvement in both performance and quality of solutions when \gls{ParDen-Sur} is applied to the aforementioned problems with the underpinning \glspl{EA} considered. However, reservoir sampling-enabled look-ahead has the most impact on the results.
The study has demonstrated the benefit of \gls{ParDen-Sur} for the portfolio optimisation hyper-parameter search problem. Future work will look to explore additional markets as well as additional problem domains in which \gls{ParDen-Sur} might be of value.

\backmatter
\bmhead{Acknowledgements}
This study was supported in part by the National Research Foundation of South Africa (Grant Number: 129541).
This study was also supported in part by the Nedbank Research Chair.

\section*{Declarations}

\subsection{Competing Interests}
The authors have no competing interests to declare. 
\subsection{Availability of data and materials} The datasets generated during and/or analysed during the current study are available on GitHub, \href{https://github.com/intelligent-systems-modelling/surrogate-assisted-moo-cvxport}{here}.

\subsection{Code availability} The source code used during the current study is available on GitHub, \href{https://github.com/intelligent-systems-modelling/surrogate-assisted-moo-cvxport}{here}.

\bibliography{sn-bibliography}


\end{document}